\documentclass[a4paper,twoside]{article}

\usepackage{epsfig}
\usepackage{subfigure}
\usepackage{calc}
\usepackage{amssymb}
\usepackage{amstext}
\usepackage{amsmath}
\usepackage{amsthm}
\usepackage{array}
\usepackage{multicol}
\usepackage{pslatex}
\usepackage{apalike}
\usepackage{microtype}
\usepackage{SCITEPRESS}     

\begin{document}

\title{Fully Convolutional Crowd Counting On Highly Congested Scenes}

 \author{\authorname{Mark Marsden, Kevin McGuinness, Suzanne Little  and Noel E. O'Connor}
 \affiliation{Insight Centre for Data Analytics, Dublin City University, Ireland}
 \email{mark.marsden@insight-centre.org}
 }


\abstract{ In this paper we advance the state-of-the-art for crowd counting in high density scenes by further exploring the idea of a fully convolutional crowd counting model introduced by \cite{zhang2016single}. Producing an accurate and robust crowd count estimator using computer vision techniques has attracted significant research interest in recent years.  Applications for crowd counting systems exist in many diverse areas including city planning, retail, and of course general public safety. Developing a highly generalised counting model that can be deployed in any surveillance scenario with any camera perspective is the key objective for research in this area. Techniques developed in the past have generally performed poorly in highly congested scenes  with several thousands of people in frame \cite{rodriguez2011density}. Our approach, influenced by the work of \cite{zhang2016single}, consists of the following contributions: (1) A training set augmentation scheme
that minimises redundancy among training
samples to improve model generalisation
and overall counting performance; (2) a deep, single column, fully convolutional network (FCN) architecture; (3) a multi-scale averaging step during inference. The developed technique can analyse images of any resolution or aspect ratio and achieves state-of-the-art counting performance on the Shanghaitech Part\_B and UCF\_CC\_50 datasets as well as competitive performance on Shanghaitech Part\_A.}

\onecolumn \maketitle \normalsize \vfill

\section{\uppercase{Introduction}}
\label{sec:introduction}

\noindent Vision based crowd size estimation, often referred to as crowd counting, has become an important topic for the computer vision community. Crowd counting algorithms attempt to produce an accurate estimation of the true number of people present in a crowded scene. A crowd count is inherently more objective than other crowd size representations (e.g. crowd density level) but is also more challenging to produce.  Accurate knowledge of the crowd size in a public space can provide valuable insight for tasks such as city planning, analysing consumer shopping patterns as well as maintaining general crowd safety. Several key challenges such as visual occlusions and high levels of variation in scene content have limited progress in this area. Techniques developed for crowd counting can also be applied to tasks from other domains such as counting bacteria or cells in microscopic images \cite{XieCount}.

\noindent \textbf{Related work.}  Existing approaches to crowd counting largely fall into two categories: counting by detection and counting by regression.

Counting by detection approaches involve training a visual object detector to find and count each person in the scene. Each human is assumed to be an individual entity that must be found. These algorithms \cite{wu2005detection,lin2001estimation,ge2009marked} are computationally demanding, requiring the image to be exhaustively analysed at multiple scales due to perspective issues, which alter the size of people in different parts of the scene. The robustness of these object detectors also suffers significantly due to visual occlusions, resulting in rapid performance degradation as a crowd becomes highly congested (i.e. several hundred people in frame).

Counting by regression techniques \cite{change2013semi,chen2012feature,lempitsky2010learning,liu2014robustness,chan2012counting} on the other hand attempt to learn a direct mapping between low-level features and the overall number of people in frame or within a frame region. Individual people are not explicitly detected or tracked in these approaches, meaning visual occlusions have less impact on counting accuracy.  While generally more computationally efficient than counting by detection methods, regression-based techniques have suffered greatly from overfitting in the past due to a lack of varied training data. To remedy this, a number of high density, high variation crowd counting datasets such as UCF\_CC\_50 \cite{idrees2013multi} and Shanghaitech \cite{zhang2016single} have emerged. Recent advancements in graphical processing unit (GPU) hardware and the availability of very large, labelled datasets such as ImageNet \cite{imagenet_cvpr09}  have resulted in deep learning approaches such as convolutional neural networks (CNN) achieving state-of-the-art performance in many computer vision tasks (image classification, face detection, object detection). Deep learning techniques have recently been applied to the task of regression-based crowd counting \cite{Zhang2015,Hu2016,zhang2016single}, resulting in a notable improvement in counting accuracy, especially for high density scenes (i.e. where there are 1000+ people in frame).

Fully convolutional networks (FCN) are a unique variation on the CNN technique where a proportionally sized feature map output is produced for a given input image rather than a classification label or regression score. FCNs have been used for a variety of tasks including semantic segmentation \cite{long2015fully} and  saliency prediction \cite{pan2016shallow}. Zhang et al. (Zhang et al., 2016) trained an FCN to
transform an image of a crowded scene into a crowd
density heatmap, which when integrated produces a
highly accurate crowd count estimate, even for very challenging scenes.  One of the key aspects of fully convolutional nets that makes the method particularly suited to crowd counting is the use of a variable size input, allowing the model to avoid the loss of detail and visual distortions typically encountered during image downsampling and reshaping.

\begin{figure*}
	\centering
	\includegraphics[width=.6\linewidth]{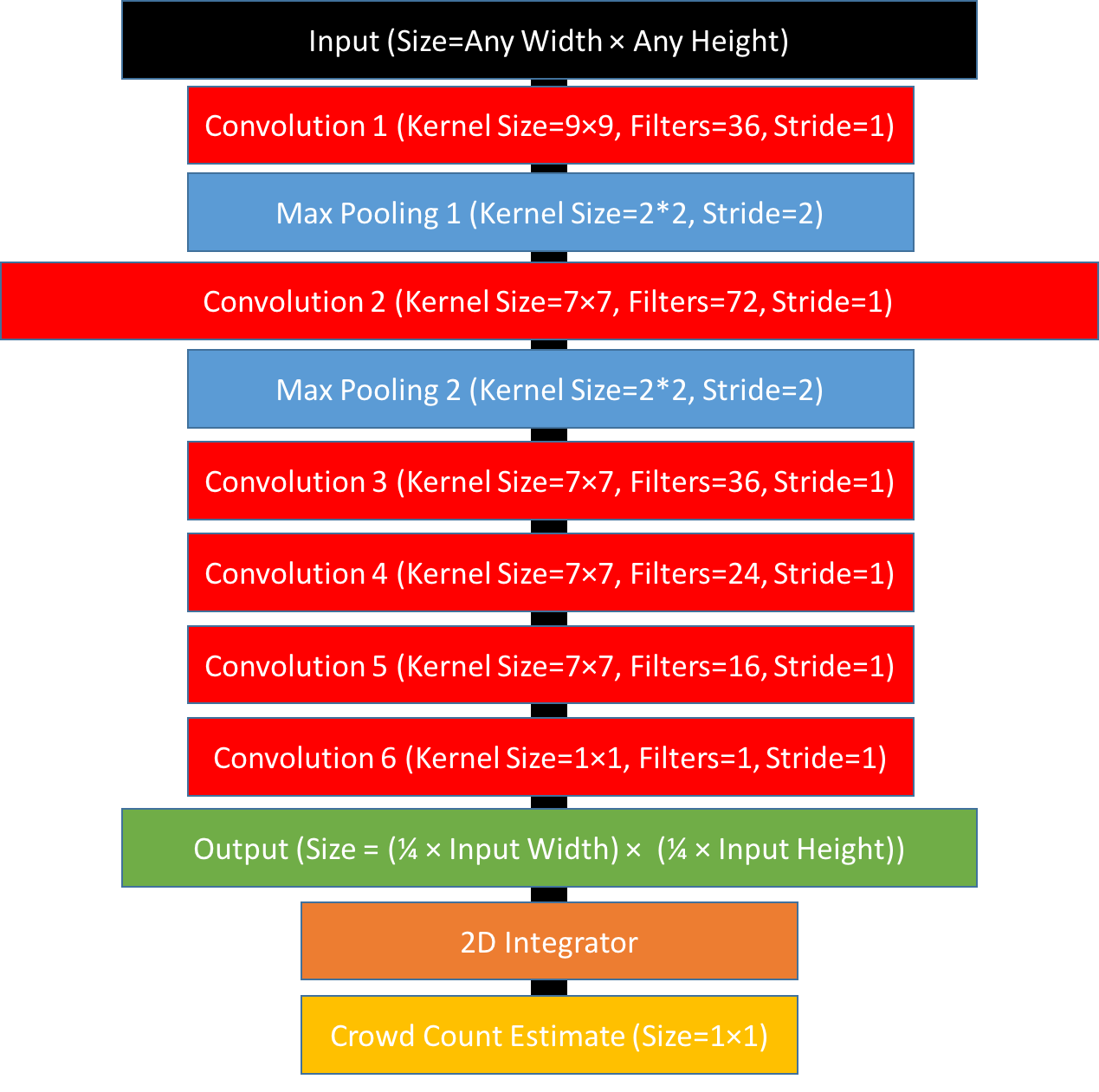}
	\caption{Fully convolutional network architecture used to perform crowd counting. Each convolutional layer is followed by a ReLU activation layer apart from "Convolution 6". A 2D integration (simply an element-wise sum in this case) is applied to the network output in order to produce the crowd count estimate value.}
    \label{arch}
\end{figure*}

\noindent \textbf{Contributions of this paper.} The core objective of this paper is to achieve highly accurate crowd counting on densely congested scenes. This study will further explore the idea of a fully convolutional crowd counting model originally introduced by \cite{zhang2016single}. The core contributions can be summarized as follows:

\begin{enumerate}
  \item  A training set augmentation scheme is proposed which  minimises redundancy among training samples in order to improve model generalisation and overall counting performance.
  
  \item A deep, single column, fully connected network is used to generate crowd density heatmaps. The greater model capacity improves the FCNs ability to learn the highly abstract, nonlinear relationships present in crowd counting datasets.
  
 \item  To overcome the scale and perspective issues that often limit the accuracy of crowd counting algorithms a given test image is fed into the network at multiple scales (e.g. original size + 80\% original size). The crowd count is estimated for each scale and the mean is taken as the overall estimate. This simple step taken during inference results in significant performance gains.
  
\end{enumerate}

\begin{table}[tbp]

\centering
\caption{Shanghaitech Part\_B validation performance using different training set augmentation schemes. Horizontal flips are used in all cases.}
\begin{tabular}{|m{3cm}|m{1cm}|m{1cm}|}

\hline
\textbf{Augmentation Scheme} &  \textbf{MAE} & \textbf{MSE} \\ \hline \hline
None & 30.5 & 47.5 \\ 
4 quadrants crops +  1 overlapping centre crop & 25.5 & 36.5 \\ 
4 quadrants crops & \textbf{24.1} & \textbf{33.5} \\ \hline 
\end{tabular}
\label{GMM_VF}
\end{table}

\section{\uppercase{A Fully Convolutional Network for Crowd Counting}}
\noindent A fully convolutional network (FCN) allows for the input images used during training and inference to be of any resolution and aspect ratio, thanks to the absense of any fully connected layers. Rather than produce a fixed size classification or regression output, FCNs generate a feature map or set of feature maps proportionally sized to the input image. This type of network can then be used for a range of tasks including image transformation and pixel wise regression/classification \cite{pan2016shallow,long2015fully}.

Zhang et al. \cite{zhang2016single} trained an FCN to transform an image of a crowded scene into a crowd density heatmap, which when integrated produces a highly accurate crowd count estimate. In order to train a network to produce this function a set of ground truth heatmap images must be generated for which the integral is equal to the pedestrian count. The head annotations found in most crowd counting datasets can be used to this end. For each of the $N$ head annotations associated
with a given training image a unit impulse is
added to the heatmap ground truth at the given location, as described in equation \ref{impulse_function} where $x_{i}$ is the position of a given head.

\begin{equation}
H(x)=\sum_{i=0}^{N}\delta(x-x_{i})
\label{impulse_function}
\end{equation}.

To convert this discrete density heatmap to a continuous function, convolution with an adaptive Gaussian kernel $G_{\sigma i}$ is applied for each head annotation \cite{zhang2016single}. The spread parameter $\sigma$ used for a given head annotation $x_{i}$ is decided based on the mean distance to the 5 nearest heads $\bar{d}_{i}$,  using equation  \ref{distance_spread}. Distance to the surrounding heads roughly correlates with proximity to the camera, producing more smoothing the closer to the camera a pedestrian is, helping us account for perspective distortion issues. The 0.3 weighting was found empirically by \cite{zhang2016single}  to produce optimal results and is maintained. This fully convolutional approach to crowd counting will form the basis of our technique.

\begin{equation}
G_{\sigma i}=0.3 * \bar{d}_{i}
\label{distance_spread}
\end{equation}

\subsection{Training Set Augmentation Scheme}
\noindent The training set generation scheme used and particularly the chosen augmentation techniques, play an important role in the strong counting accuracy achieved by our method. Most crowd counting datasets consist of only a few hundred images, making augmentation an essential step. Taking several image crops to increase training set size and variation is a common augmentation technique used in computer vision.  While it is perfectly acceptable to allow these crops to overlap for image recognition tasks, pixel-wise tasks can potentially overfit when the network is continually exposed to a given set of pixels during training. Therefore our augmentation scheme is developed to ensure there is no such redundancy. For each training set image the four image quadrants as well as their horizontal flips are taken as training samples, ensuring no overlap.
In order to validate this augmentation scheme the Shanghaitech Part\_B training set is further split into training and validation subsets using a 9:1 ratio. Table \ref{GMM_VF} highlights the difference in validation performance when our model is trained on a dataset with and without overlapping crops. Both runs are trained from scratch using the same network architecture. This simple change results in a notable improvement in counting accuracy, despite the reduction in overall training set size.

\subsection{FCN Architecture}
\noindent Processing high resolution images (e.g. $1000 \times 1000$ pixels) using a fully connected network presents certain challenges and constraints, particularly in terms of memory usage on GPU hardware. We are limited in the number of convolutional kernels and layers (i.e. model capacity) our FCN can have. Therefore we must attempt to design the best possible FCN architecture capable of processing high resolution images such as those in the UCF\_CC\_50 dataset. An Nvidia GTX 970 card with 4GB of VRAM was used for our experiments. With these constraints in mind we designed a 6 layer, single column FCN as illustrated in figure \ref{arch}. This network contains just 315,000 parameters, thanks largely to the absence of any fully connected layers.
Rectified linear unit (ReLU) activations are applied after each convolutional layer apart from the last. $1 \times 1$ convolutions are used in the final layer to produce the single channel crowd density heatmap. This density heatmap is then fed into a 2D integrator (simply an element-wise sum in this case) to produce the crowd count estimate. The network is optimised in a single training run using stochastic gradient descent and backpropagation.  We chose to minimise the Euclidean distance between the produced density heatmap and the ground truth heatmap. This loss function is fully defined as follows:

\begin{equation}
L(\Theta )=\frac{1}{2N}\sum_{i=1}^{N}\left \| F(X_{i};\Theta)-F_{i} \right \|_{2}^{2},
\end{equation}

\noindent where  $\Theta $ is the set of  network parameters to optimise, \textit{N}  is the batch size, \textit{X\_{i}} is the $i^{th}$  batch image and F\_{i} is the corresponding ground truth density heatmap. $F(X_{i};\Theta)$ is the estimated heatmap for a given batch image \textit{X\_{i}}.

\begin{table}[t]
\caption{Shanghaitech Part\_B validation performance using different network architectures.}
\centering
\begin{tabular}{|m{4cm}|m{1cm}|m{1cm}|}

\hline
\textbf{Network Architecture} & \textbf{MAE} & \textbf{MSE} \\ \hline \hline
Proposed & \textbf{24.1} & \textbf{33.5} \\ 
Multi-Column FCN \cite{zhang2016single} & 25.5 & 36.5 \\ \hline
\end{tabular}
\label{val_net}
\end{table}

Table \ref{val_net} presents the difference in Shanghaitech Part\_B validation performance  when our high capacity architecture is used over a shallower multi-column FCN architecture \cite{zhang2016single}. All hyperparamaters including the training set augmentation scheme are kept identical for both runs. We can see a clear improvement in performance when our deeper single column architecture is used.

\begin{table*}[t]
\caption{Shanghaitech Part\_B validation performance using different mutli-scale averaging inference schemes.}
\centering
\begin{tabular}{|m{9cm}|m{1.5cm}|m{1.5cm}|}
\hline
\textbf{Multi-scale Averaging Scheme} & \textbf{MAE} & \textbf{MSE} \\ \hline \hline
1) None & 24.1 & 33.5 \\
2) MeanCount(Original Size, 80\% Original Size) & \textbf{22.1} & \textbf{31.5} \\ 
3) MeanCount(Original, 80\% Original Size, 70\% Original Size) & 24.6 & 34.1 \\ 
4) MeanCount(Original, 80\% Original Size, 70\% Original Size, 60\% Original Size) & 25.2 & 34.8 \\ \hline
\end{tabular}
\label{val_scale}
\end{table*}

\subsection{Multi-Scale Averaging During Inference}
\noindent Scale and perspective  issues often limit  the performance of crowd counting algorithms. A top down camera perspective is ideal for this task but cannot be guaranteed in real world settings.  In most CCTV scenarios foreground pedestrians are much larger than those in the background, who may only occupy a few pixels. As FCNs allow for a variable size input image, we can easily resize a given test image before feeding it into the network and estimating the crowd size. A scaled down version may result in more accurate crowd counting in certain scene regions than the original.  Therefore in order to overcome these issues a given test image is fed into the network at multiple scales (e.g. original size + 80\% original size). The crowd count is estimated for each scale and the mean is taken as the overall estimate. Table \ref{val_scale} shows the validation performance of several multi-scale averaging schemes. The same training and validation subsets are used as before.  Scheme 2 performs best and is thus used for all further experiments.

\begin{figure*}[t]
	\centering
	\includegraphics[width=0.26\textwidth]{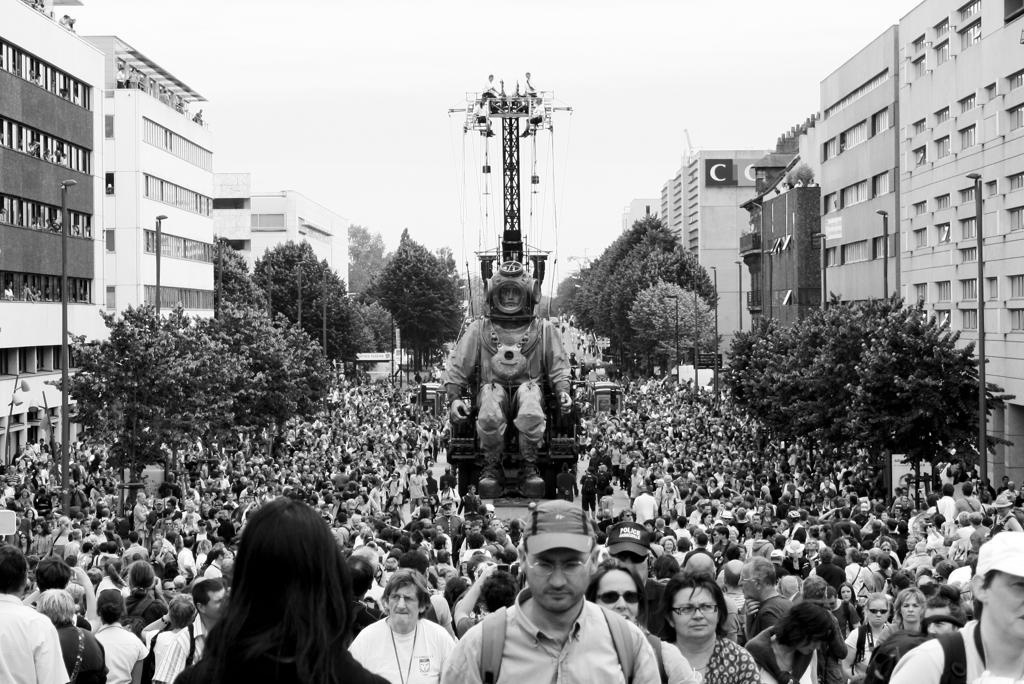}
    \includegraphics[width=0.26\textwidth]{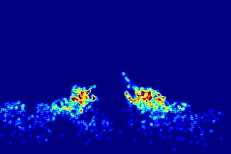}
    \includegraphics[width=0.26\textwidth]{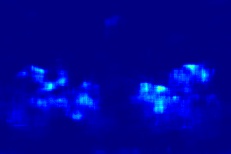}
   	\includegraphics[width=0.26\textwidth]{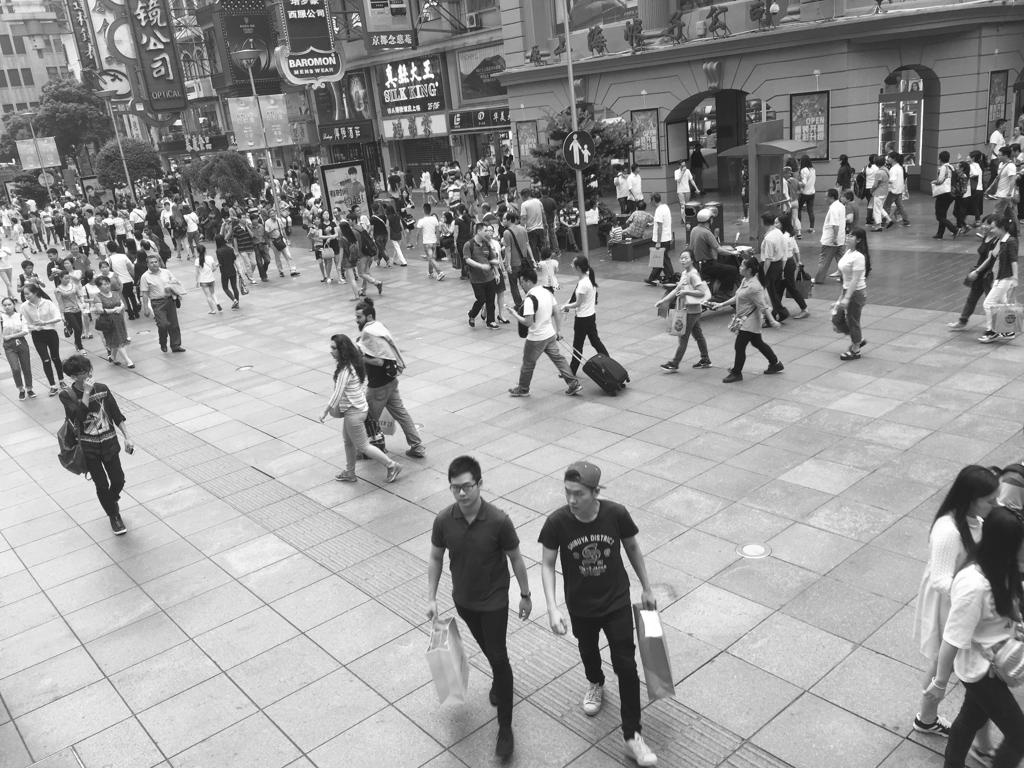}
    \includegraphics[width=0.26\textwidth]{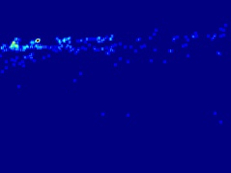}
    \includegraphics[width=0.26\textwidth]{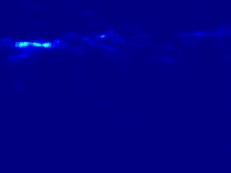}

	\caption{\textbf{Top:} Shanghaitech Part\_A test image, Ground Truth Heatmap, Estimated Heatmap. True Count=1326, Estimated Count=1138. \textbf{Bottom:} Shanghaitech Part\_B test image, Ground Truth Heatmap, Estimated Heatmap. True Count=240, Estimated Count=242.}
    \label{SHT_action}
\end{figure*}

\section{\uppercase{Experiments}}
\noindent



\noindent The performance of our fully convolutional crowd counting technique is evaluated on three crowd counting benchmarks from two datasets. These benchmarks vary greatly in terms of congestion level and scene content. Our model achieves very strong crowd counting performance, particularly on images of high density scenes with several thousand people in frame. The Caffe framework \cite{jia2014caffe} and it's Python wrapper are used to train and deploy our model.  Both mean absolute error (MAE) and mean squared error (MAE) are used to compare crowd counting performance on all datasets. These two metrics are defined as follows:

\begin{equation}
\text{MAE}=\frac{1}{N}\sum_{i=1}^{N}\left |z_i-\check{z}_i  \right |, 
\end{equation}
\begin{equation}
\text{MSE}=\sqrt{\frac{1}{N}\sum_{i=1}^{N}( z_i-\check{z}_i  )^{2} },
\end{equation}

\noindent where \textit{N} is the number of test images, \textit{z\_{i}}  is the actual number of people in the $i^{th}$ image and ˆ
\textit{$\check{z}_i $} is the estimated number of people in the $i^{th}$ image. MAE indicates the accuracy of the estimates while MSE corresponds to the robustness of the estimates.

\subsection{Shanghaitech Dataset}
\noindent The Shanghaitech dataset \cite{zhang2016single} contains 1198 images of crowded scenes with a total of 330,165 head annotations included. This dataset is split into two parts; Part\_A contains images of high density scenes (up to 3000 people) taken from the internet while Part\_B consists of medium density crowd images (up to 600 people) taken in the streets of Shanghai. Each part consists of a respective training and test set. The performance of the proposed approach is evaluated on each part separately.

The redundancy minimising augmentation approach discussed in section 2.1 is used for both parts.  The network is trained from scratch in a single run for $2e^{6}$ iterations using a base learning rate of $1e^{-6}$, with the learning rate decreased by a factor of 10 after $1e^{6}$ iterations. Gaussian weight initialisation with a standard deviation of 0.01 is used as well as a weight decay of 0.0005 and a momentum of 0.9. Due to memory limitations and the high image resolution of the Shanghaitech dataset a batch size of 1 is used during training. During testing the proposed multi-scale averaging is applied, using scheme 2 from table \ref{val_scale}.

Our method is compared to the existing approaches in table \ref{perf_SHT} and achieves state-of-the-art performance on Shanghaitech Part\_B, improving MAE by 10\% and MSE by 19\%. Competitive performance is also achieved on Shanghaitech Part\_A, with an MSE near identical to the state-of-the-art produced.   Figure \ref{SHT_action} shows our technique in action on images from this dataset.

\begin{table*}[t]
\caption{Comparing the performance of different crowd counting approaches on the Shanghaitech dataset.}
\centering
\begin{tabular}{|m{3cm}|l|l|l|l|}
\hline
                                   & \multicolumn{2}{c}{\textbf{Part\_A}} \vline        &  \multicolumn{2}{c}{\textbf{Part\_B}} \vline    \\ \hline 
\textbf{Method}                             & \textbf{MAE}            & \textbf{MSE}           & \textbf{MAE}           & \textbf{MSE}           \\ \hline \hline
 \cite{zhang2016single}                 & \textbf{110.2}          & \textbf{173.2}          & 26.4          & 41.3          \\ \hline
\cite{Zhang2015} & 181.8          & 277.7          & 32.0          & 49.8          \\ \hline
Proposed Approach                       & 126.5 & 173.5 & \textbf{23.76} & \textbf{33.12} \\ \hline
\end{tabular}
\label{perf_SHT}
\end{table*}

\subsection{UCF\_CC\_50 Dataset}
\begin{figure*}[t]
\centering

\includegraphics[width=0.27\textwidth]{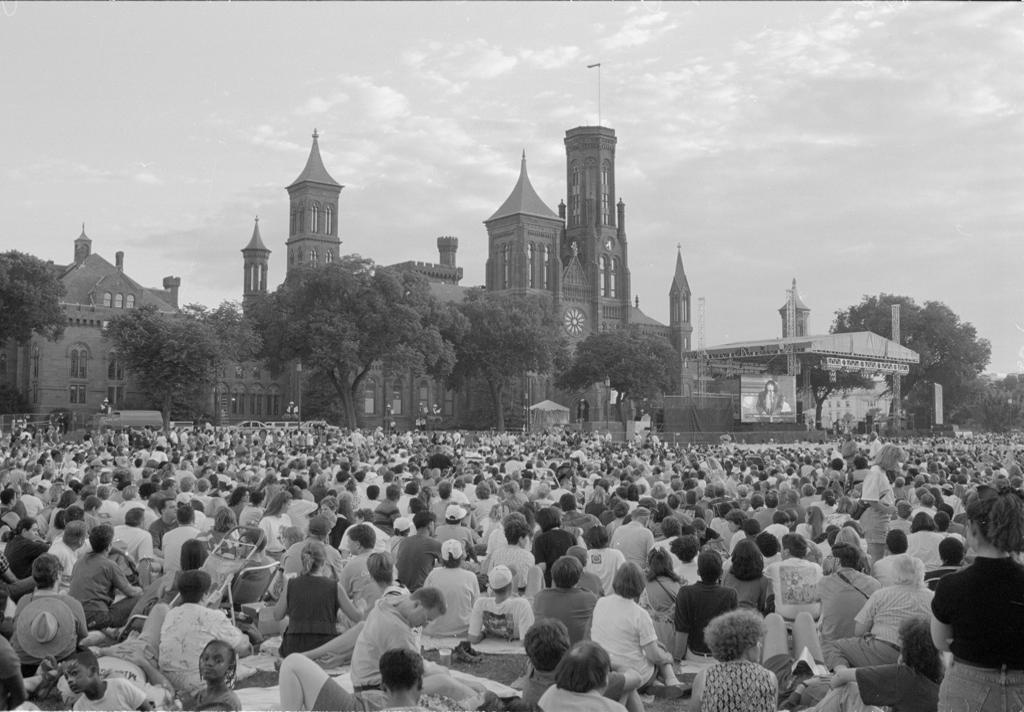}
    \includegraphics[width=0.27\textwidth]{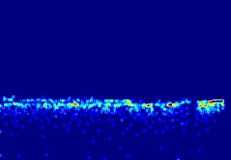}
    \includegraphics[width=0.27\textwidth]{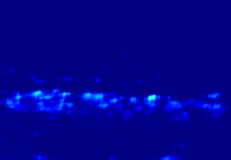}
    
	\caption{UCF\_CC\_50 test image, Ground Truth Heatmap, Estimated Heatmap. True Count=1544, Estimated Count=1566.}
    \label{UCF_action}
\end{figure*}

\noindent The UCF\_CC\_50 dataset \cite{idrees2013multi} contains 50 highly challenging crowd images taken from the Internet. The number of pedestrians present in a frame ranges between 94 and 4500. Following convention \cite{idrees2013multi} a 5-fold cross validation is performed on this dataset. The same augmentation process, training hyperparamaters and multi-scale averaging scheme are used as for the Shanghaitech dataset. Again due to memory limitations a batch size of 1 is used. Table \ref{perf_UCF} compares our technique with the existing approaches, with our method improving the state-of-the-art for MAE and MSE by 11\% and 13\% respectively. Figure \ref{UCF_action} shows our technique in action on an image from this dataset.

\begin{table}[t]
\caption{Comparing performance of different crowd counting approaches on the UCF\_CC\_50 dataset.}
\centering
\begin{tabular}{|m{4cm}|l|l|l|}
\hline
\textbf{Method}                             & \textbf{MAE}            & \textbf{MSE}                  \\ \hline \hline
 \cite{rodriguez2011density}                 & 655.7          & 697.8       \\ \hline
 \cite{lempitsky2010learning}                 & 493.4         & 487.1      \\ \hline
 \cite{idrees2013multi}                 & 419.5        & 541.6       \\ \hline
 \cite{zhang2016single}                 & 377.6          & 509.1       \\ \hline
  \cite{Hu2016}                 & 431.5          & 438.5       \\ \hline
 \cite{Zhang2015} & 467.0          & 498.6      \\ \hline
Our Approach                       & \textbf{338.6} & \textbf{424.5}  \\ \hline
\end{tabular}
\label{perf_UCF}
\end{table}

\begin{figure}[t]
	\centering
	\includegraphics[width=0.5\textwidth]{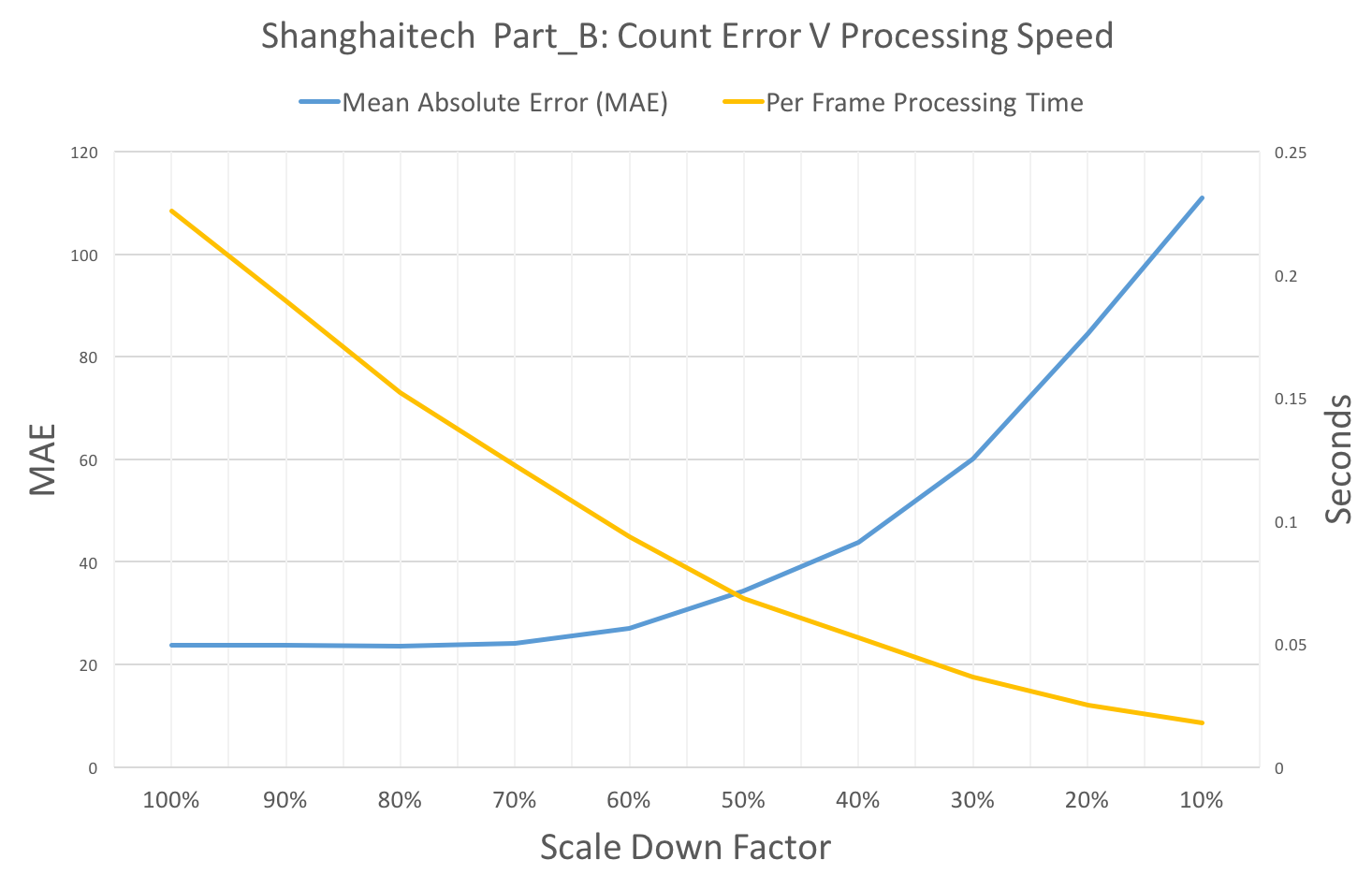}

	\caption{Comparing count error and processing speed as we scale down the image resolution on Shanghaitech Part\_B.}
    \label{tradeoff}
\end{figure}

\subsection{Cross Dataset Performance}
\noindent In order to investigate the generalisation potential of our technique, we performed a number of cross dataset experiments. In each experiment we take a model trained on a specific dataset as the "source" domain and then evaluate MAE and MSE performance on another unseen dataset or "target" domain. The results of these experiments are shown in Table \ref{trans}. Superior performance is achieved when our source and target domain both contain images of a similar density level (Shanghaitech Part\_B $=>$ Shanghaitech Part\_A, Shanghaitech Part\_A $=>$ UCF\_CC\_50). On the other hand very poor performance is achieved when our source domain contains significantly higher density images than the target (UCF\_CC\_50 $=>$  Shanghaitech Part\_A). Therefore a model used for real world deployment must be trained on an appropriately large and varied training set.

\begin{table*}[t]
\caption{Cross dataset performance of our method. The percentage increases in MAE and MSE are highlighted.}
\centering
\begin{tabular}{|l|l|l|l|}
\hline
\textbf{Source Domain}                   & \textbf{Target Domain}   & \textbf{MAE}   & \textbf{MSE}  \\ \hline
Shanghaitech\_B                 & Shanghaitech\_A & 191(+52\%)   & 337.5(+94\%)  \\ \hline
UCF\_CC\_50                     & Shanghaitech\_A & 269(+116\%) & 359.5(107\%)  \\ \hline
Shanghaitech\_A                & Shanghaitech\_B & 68(+189\%) & 100.5(+200\%) \\ \hline
UCF\_CC\_50                     & Shanghaitech\_B & 165(+614\%) & 215(+540\%) \\ \hline
Shanghaitech\_A                 & UCF\_CC\_50     & 473(+40\%)   & 680(+50\%) \\ \hline
Shanghaitech\_B                 & UCF\_CC\_50     & 699(+100\%)   & 866 (+105\%) \\ \hline

\end{tabular}
\label{trans}
\end{table*}

\subsection{Trade-off between Computation Speed and Counting Accuracy}
\noindent The ability of a fully convolutional network to process images of any resolution is one of the key reasons behind the strong counting performance achieved by our method. However, analysing such high resolution images results in high memory consumption and slower processing speed during inference. Therefore we want to investigate to what degree image resolution can be reduced during test time before we see significant performance degradation. Doing so we can find the best possible trade-off between computation speed and accuracy. The Shanghaitech Part\_B dataset is used for this experiment. Test images are scaled down to a given percentage of their original size with aspect ratio maintained. The results of these experiments are presented in figure \ref{tradeoff}. Surprisingly we do not see the error increase significantly until we reduce the image size to 50\%. However with this 50\% resizing applied the processing speed is increased by a factor of 4. In deployment scenarios this type of downsampling can be applied in order to analyse real-time video without a major loss of accuracy.

\section{\uppercase{Conclusion}}
\noindent In this paper we have proposed a deep, fully convolutional crowd counting model that can perform highly accurate single image crowd counting in almost any surveillance scenario. Our model achieves state-of-the-art performance on both the Shaghaitech Part\_B and  UCF\_CC\_50 datasets as well as competitive performance on the Shaghaitech Part\_A dataset. Images of any resolution and aspect ratio can be analysed.  The developed approach also performs well even with significant image downsampling applied at test time. Future work in this area will look to extend our network to also perform other pixel-wise tasks such as crowd segmentation in order to exploit the inter-task correlations present.

\section{\uppercase{Acknowledgements}}
\noindent  This publication has emanated from research conducted with the financial support of Science Foundation Ireland (SFI) under grant number SFI/12/RC/2289


\vfill
\bibliographystyle{apalike}
{\small
\bibliography{example}}

\end{document}